\documentclass[sn-mathphys]{sn-jnl} 
\jyear{2023}

\theoremstyle{thmstyleone}

\theoremstyle{thmstyletwo}

\usepackage{comment}
\usepackage[utf8]{inputenc}          
\usepackage[T1]{fontenc}              
\usepackage[USenglish]{babel}         
\usepackage{array}
\usepackage{tabularx}
\usepackage{caption}
\usepackage{graphicx}
\usepackage{mathtools}
\usepackage{setspace}
\usepackage{graphicx}
\usepackage{hyperref}

\usepackage{arabtex}
\usepackage{utf8}
\usepackage{graphicx}
\usepackage{diagbox}
\usepackage{algorithm}
\begin{document}

\title{AraSpell: A Deep Learning Approach for Arabic Spelling Correction}

\author[]{\fnm{Mahmoud} \sur{Salhab}
%\dgr{MSc, PhD}
}%\email{mahmoud.salhab@lau.edu}
\author[]{\fnm{Faisal} \sur{Abu-Khzam}%\dgr{MSc, PhD}
}%\email{faisal.abukhzam@lau.edu.lb}
\affil[]{\centering\orgdiv{Department of Computer Science and Mathematics}\\ \orgname{Lebanese American University}\\ \orgaddress{\city{Beirut} \postcode{1102 2801}, \country{Lebanon}}}
%\affil[2]{\orgdiv{Fachbereich~4 -- Abteilung Informatikwissenschaften}, \orgname{Universit\"at Trier}, \orgaddress{\street{Universit\"atsring 15}, \city{Trier}, \postcode{54286}, \state{Rheinland-Pfalz}, \country{Germany}}}

%

\thispagestyle{empty}

\abstract{Spelling correction is the task of identifying spelling mistakes, typos, and grammatical mistakes in a given text and correcting them according to their context and grammatical structure. This work introduces "AraSpell," a framework for Arabic spelling correction using different seq2seq model architectures such as Recurrent Neural Network (RNN) and Transformer with artificial data generation for error injection, trained on more than 6.9 Million Arabic sentences.
Thorough experimental studies provide empirical evidence of the effectiveness of the proposed approach, which achieved 4.8\% and 1.11\% word error rate (WER) and character error rate (CER), respectively, in comparison with labeled data of 29.72\% WER and 5.03\% CER. Our approach achieved 2.9\% CER and 10.65\% WER in comparison with labeled data of 10.02\% CER and 50.94\% WER. Both of these results are obtained on a test set of 100K sentences.
}
\keywords{Arabic spelling correction, Seq2Seq, Deep Learning, Neural Networks.}
\footnote{Available on GitHub at: https://github.com/msalhab96/AraSpell}

\maketitle              % typeset the header of the contribution

\section{Introduction}

In recent years, the advances of Natural Language Processing (NLP) methods, often applied to machine translation, speech recognition, question answering, 
%, text summarization 
among others, have led to a wide adoption of NLP solutions in various fields like health care \cite{nlphealthcare}, manufacturing \cite{Prakash2019-oy}, finance \cite{xing2018a}...etc, mostly focusing on English, with scarce attention on the Arabic language. 

Arabic is the 4\textsuperscript{th} most-used language on the internet, and it is one of the six official languages of the United Nations. It is spoken by people across 22 countries \cite{article, 6841973}. Arabic can be categorized into three main variants: (i) Classical Arabic (CA), which is mainly used
in ancient and theological texts but is still understood due to its use in the Holy
Quran, (ii) the Modern Standard Arabic (MSA), which is a modernized and simplified version of CA, and (iii) Dialectal/colloquial Arabic (DA), where each region has
its own dialect. The most widely-understood variation among Arabic speakers is MSA due to its wide adoption in education, media, and formal communication across the different Arabic-speaking countries \cite{ALAYYOUB2018522}.

Spelling correction is the task of identifying the incorrect words in a sentence and then correcting them according to certain criteria. Spelling errors can greatly impact the output of sophisticated natural language processing (NLP) and natural language understanding (NLU) models, and it can greatly degrade the accuracy of these models. For example, Napoles et al. \cite{inproceedings} show how error injection could badly affect the performance of a dependency parsing system.

Spelling correction modules are vital components for many real-world NLP applications such as Optical Character Recognition (OCR) systems where spelling correction is employed in the post-processing phase to possibly correct any misspelled word \cite{DBLP:journals/corr/abs-1204-0191}. Moreover, it is used for search query correction to ensure the quality of search-based systems \cite{Rachidi_arabicuser}. 

Most of the recent work in the Arabic spelling correction field has relied on traditional approaches such as statistical and rule-based approaches, that usually consist of two cascaded systems, namely error detection, and correction, where a word would pass to the correction system if and only if an error gets detected \cite{Moslem2020ArabiscCN}.

This paper presents an end-to-end framework for automatic Arabic spelling correction using various Seq2Seq models trained on more than 6.9 million sentences, starting from a large number of correct sentences and introducing various levels of errors via a novel error-injection scheme. The paper is structured as follows. Literature review is presented in the next section, followed by a section describing the proposed model, and the data used. Section 4 presents the experiments and their results and we conclude in Section 5 with a summary and potential future directions.

\section{Literature Review}

Automatic spelling correction received considerable attention from the research community. Kukich \cite{Kukich1992TechniquesFA} divides the process of spelling correction into three steps: (i) error detection, (ii) correction candidate generation, (iii) candidate ranking. A hierarchical character tagger model presented in \cite{Gao2021HierarchicalCT} for short text spelling error correction, where they used a pre-trained language model at the character level as a text encoder, and then predict character-level edits to transform the original text into its error-free form with a much smaller label space, and for decoding they proposed a hierarchical multi-task approach to alleviate the issue of long-tail label distribution without introducing extra model parameters. A word vector/conditional random field (CRF)-based detector was proposed in \cite{Wang2015WordVR} to detect Chinese spelling errors and used a language model for re-scoring.

The Long Short-Term Memory 
model (LSTM) proposed in \cite{8904218} encodes the input word at the character level, which also uses word and POS tag contexts as features for Indonesian text. Similarly, \cite{Kinaci2018SpellingCU} trained a recurrent neural network with dictionary words. For a given misspelled word, they retrieve a candidate list from that dictionary word, then the list gets expanded using a character-level bi-gram model and the trained model.

For Arabic, a set of correction rules is proposed in \cite{Nawar2014FastAR} such that these rules get ranked by calculating their probability before they get applied to the input text. In \cite{rulebased1} Shaalan et al. proposed a tool capable of recognizing and suggesting correction of ill-formed input for common spelling errors. It  is  composed  basically  of  Arabic  morphological  analyzer, lexicon, spelling  checker, and spelling corrector. 

A system of two components introduced in \cite{6193415}: one
to retrieve candidates for misspelled words using Damerau–Levenshtein, and another one
to correct spelling errors using A* lattice search and 3-grams language model. 
Similarly, a two-stage system was used in \cite{paper1}, where first 
a misspelled word is detected via a morphological analysis system. The detected word is classified as incorrect if it does not have analysis, and the authors used different correction candidate generation and then used a Naive-Bayes classifier to pick the most likely to be the correction for the incorrect word. Moreover, \cite{Attia2016-qm} used a look-up table and character-level language model for misspelling detection and then used candidate generation and ranking. The candidate then get selected using a language model.

Shaalan, Khaled \cite{spellchecking1} proposed a look-up table and language modeling for error detection followed by a candidate generation that uses edit distance measured by Levenshtein distance from the misspelled word, then a noisy channel model trained on one-billion words with knowledge-based rules to assign scores to the candidate corrections and choose the best correction independent of the context. 

A neural network model using bidirectional long short-term memory (bi-LSTM) with a polynomial network (PN) for error detection is proposed in \cite{10.1145/3373266}. In \cite{Moslem2020ArabiscCN}, Moslem et al. introduced a many-to-one neural network-based context-sensitive spelling checking and correction model, where they modeled the words that come both before and after the word to be corrected as the conditional context in language
model predictions. Finally, Abandah et al. \cite{abandah} used stacked Long Short-Term Memory (LSTM) modules for common soft Arabic spelling errors correction with stochastic error injection for a limited number of characters to capture limited frequent mistakes.

\section{Methodology}

Automatic spelling correction systems detect a spelling error and propose a set of candidates for correction. Building such a system using traditional approaches - such as rule-based approaches - requires intensive work to build and debug. The advantage of deep learning stems from its ability to discover the underlying patterns in the data. Despite the fact that deep learning based systems outperform traditional systems in many fields, such systems require an intensive amount of labeled data to generalize well and to be able to be used in production.
In reality, having labeled data is rare, and the manual data labeling process is very hard, costly, and requires domain experts to do. Given the fact that unlabeled data is hugely available, the ability to do auto-labeling 
%where we get the labels from unlabeled data 
would have a great advantage to building automatic spelling correction systems.
%such systems. 
If we are confident that the unlabeled data from an initial source is accurate and trustworthy, we can build a labeled corpus as we describe in \ref{sec:data} then train the model on that data. 

\begin{figure}[htb!]
  \includegraphics[width=\linewidth]{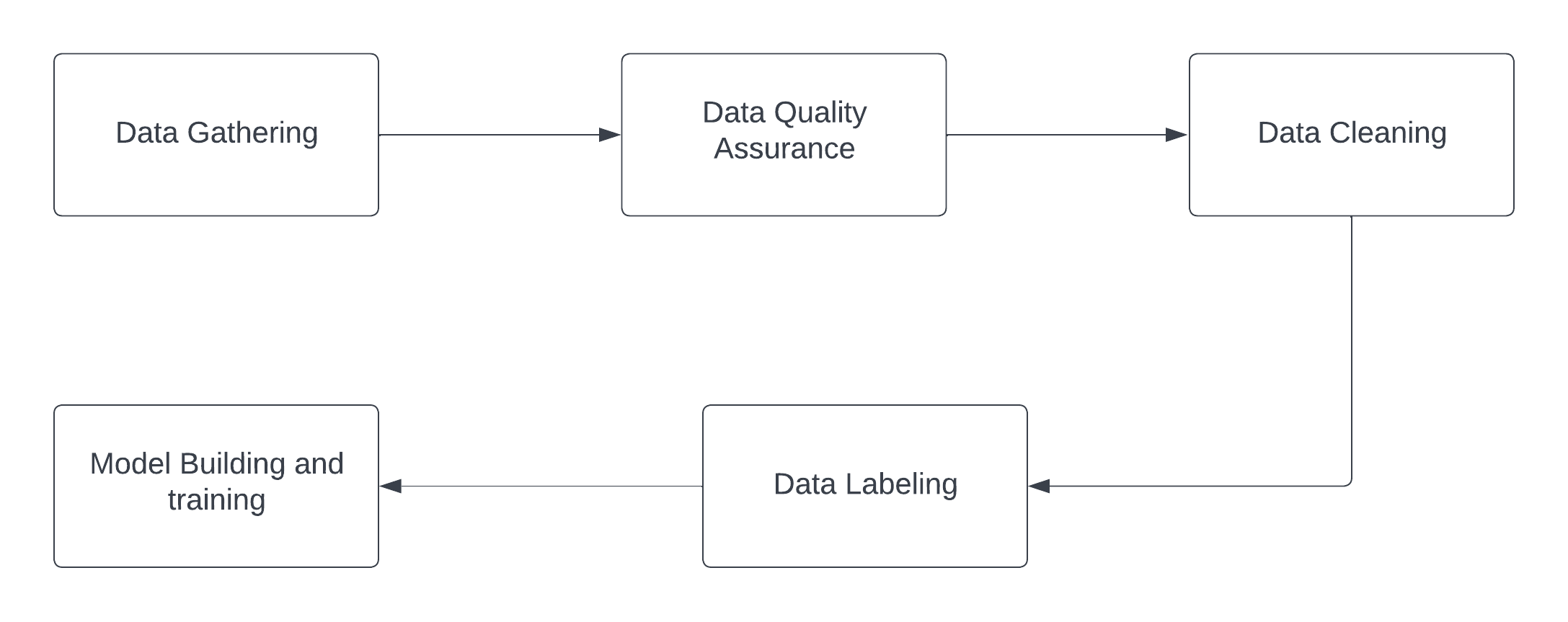}
  \caption{An overview of AraSpell project structure.}\label{fig:blocks}
\end{figure}

Figure \ref{fig:blocks} shows an overview on how our approach is structured. Our toolchain starts with data gathering, followed by data quality assurance, data cleaning and data labeling (in the specified order). Lastly, Modeling is applied where we build and train different Seq2Seq model architectures.

We now present the details of our proposed approach to solving Arabic spelling correction using various Seq2Seq models and a new self-labeling method. 

\subsection{Data}
\label{sec:data}
For this work we used the Arabic Wikipedia 2021 dump\footnote{The data set is available on Kaggle here https://www.kaggle.com/datasets/z3rocool/arabic-wikipedia-dump-2021}, the data set contains 711230 articles in various domains.

\subsubsection{Data Preprocessing}

The data comes in the form of articles. Some articles may contain different languages, so the can have encoding and formatting issues as well as invalid symbols, and might have different lengths.

Given a set of articles $ \boldsymbol{A} = \{a_1, a_2, \dots, a_H\}$ and set of ordered transformation functions $\mathcal{F}=(f_1, f_2, \dots, f_T)$ applied on each article in $\boldsymbol{A}$ resulting a set of clean lines $\boldsymbol{L}$.

Where $\boldsymbol{L}=\bigcup\limits_{h=1}^{H}f_1(f_2(\dots f_T(a_h)\dots))$ such that each line in $\boldsymbol{L}$ has a reasonable length, contains valid characters only, contains correct and valid Modern Standard Arabic (MSA) words.
The transformation functions built and used can be categorized into two groups:
\begin{enumerate}
    \item Cleaning, cleansing, and mapping: The articles are lengthy, vary in length, contain Arabic diacritization, some characters in the articles represented in various uni-codes, and lastly many articles contain non-Arabic content, for that each article split down into short paragraphs, Arabic diacritization omitted, any character repeated more than twice truncated into two characters only, characters normalized as table \ref{table:table1} shows, and only valid Arabic characters are kept.

    \item Filtering: We filtered the lines by dropping any line that either contains numbers, contains a floating character -floating character is a word in the sentence that has a length of one-, contains any citation content, contains more than two unique words -unique words are the words that appear only once in the corpus-, contains less than 3 or more than 20 words, or the number of characters in the line is less than 15 or more than 128 characters.

\end{enumerate}

\setcode{utf8}
\begin{table}[htb!]
\centering
\resizebox{\columnwidth}{!}{%
\begin{tabular}{|ccc|ll}
\cline{1-3}
\multicolumn{1}{|c|}{Arabic Character} & \multicolumn{1}{c|}{Original Unicode} & Normalized to                       &  &  \\ \cline{1-3}
\RL{لا}               & \textbackslash{}ufefb                 & \textbackslash{}u644\textbackslash{}u627 &  &  \\
\RL{لأ}               & \textbackslash{}ufef7                 & \textbackslash{}u644\textbackslash{}u623 &  &  \\
\RL{لإ}               & \textbackslash{}ufef9                 & \textbackslash{}u644\textbackslash{}u625 &  &  \\
\RL{لآ}               & \textbackslash{}ufef5                 & \textbackslash{}u644\textbackslash{}u622 &  &  \\ \cline{1-3}
\end{tabular}%
}
\caption{Arabic normalized uni-codes }
\label{table:table1}
\end{table}

\subsubsection{Artificial Data Generation}\label{datageneration}

For the task of spelling correction, as with many supervised learning tasks, a pair of correct (target) and incorrect (input) sequences are required. In order to generate the incorrect sequences for the correct ones, we propose an error injection method that takes a clean sequence and injects noise into it.

Given a set of cleaned lines $\boldsymbol{L} = \{l_1, l_2, \dots, l_N\}$ our objective is to find $\boldsymbol{\hat{L}} = \{\hat{l}_1, \hat{l}_2, \dots, \hat{l}_N\} $ such that the $\hat{l}_n$ is the corrupted/distorted version of $l_n$.
For a given clean line $l_n$, let G be a set of predefined error injection operations such that $G = \{g_1, g_2, \dots, g_M\}$ with:

\begin{equation}
\hat{l}\textsubscript{n} = \mathcal{G}_n(l\textsubscript{n})
\end{equation}
\begin{equation}
\mathcal{G}_n(l\textsubscript{n}) = g\textsubscript{n,1}(g\textsubscript{n,2}(\dots g\textsubscript{n,$\phi_n$}(l\textsubscript{n})\dots))\label{ereq}
\end{equation} 
\begin{equation}
g\textsubscript{n,$\phi_n$} \sim unif(G)
\end{equation} 
\begin{equation}
\phi\textsubscript{n} = \lfloor J_n . \psi \rfloor
\end{equation} 
\newline
Where $\phi_n$ is the number of error injection operations of the n\textsuperscript{th} line, $J_n$ is the length of the n\textsuperscript{th} line, and $\psi$ is the corruption/distortion ratio or Character Error Rate (CER).
\vspace{5pt}
\newline
Any error injection operation in G could work under one of the below modes:

\begin{itemize}
\item Insertion: a random valid character gets inserted randomly at a random position in the sentence.

\item Deletion: a random character or pattern in the sentence gets removed. 

\item Substitution: a random character in the sentence gets replaced by either a random character or one of its keyboard key neighbors.

\item Transposition of two adjacent characters: two adjacent characters get swapped randomly.

\item Mapping: a certain predefined pattern in the sentence gets replaced by one of the predefined targets. 
\end{itemize}

\subsection{Proposed Models}

Given a set of examples $\{(x_1, y_1), (x_2, y_2), \dots, (x_N, y_N)\}$ where x\textsubscript{n} and y\textsubscript{n} are the n\textsuperscript{th} corrupted text, and the clean/original text respectively. Let x\textsubscript{n} = $\{<SOS>, x_{n}^1, \dots, x_{n}^K, <EOS>\}$ the input sequence, let y\textsubscript{n} = $\{<SOS>, y_{n}^1, \dots, y_{n}^L, <EOS>\}$ the output sequence, and let the vocabulary $\mathbb{V}$ = $\{aleph, baa', \dots, yaa', <space>\}$, $x_{n}^k$, $y_{n}^l\in \mathbb{V}$, where SOS is the start-of-sentence token, and EOS is the end-of-sentence-token.
We want to model each output character $y_{n}^l$ as a conditional distribution over the previous characters and the input $x_{n}$ using the chain rule:

\begin{equation}
\prod_{l=1}^{L} P(y\textsubscript{n}\textsuperscript{l}\mid x\textsubscript{n}, y\textsubscript{n}\textsuperscript{0:l-1})\label{condeq}
\end{equation}

As in machine translation, in the task of spelling correction, the model's input sequence $x_{n}$ usually does not match the output sequence $y_{n}$: $\mid x_{n}\mid < \mid y_{n}\mid$ when the number of deletions is higher than the number of insertions, $\mid x_{n}\mid > \mid y_{n}\mid$ when the number of insertions is higher than the number of deletions, or $\mid x_{n}\mid = \mid y_{n}\mid$ in case of swapping, replacement, or when the number of insertions equals the number of deletions. We shall therefore model the conditional distribution in \eqref{condeq} using Seq2Seq encoder-decoder. For that we conducted a wide range of experiments on different Seq2Seq model architectures such as RNN-based encoder-decoder and Transformers.

\subsubsection{Attentional vanilla Seq2Seq using RNN}\label{vanilla}
The attentional vanilla Sequence-to-Sequence RNN model consists of two sub-modules: encoder and decoder with attention.
The encoder takes the input sequence $\boldsymbol{X}$ and transforms it into a high-level representation $\boldsymbol{h} = (h_1, \dots, h_K)$, while the decoder with attention takes $\boldsymbol{h}$, the previously predicted character, and the last hidden state and produces a probability distribution over character sequences:

\begin{equation}
\boldsymbol{h} = Encoder(\boldsymbol{X})\label{encodereq}
\end{equation}

\begin{equation}
P(y\textsubscript{n}\mid \boldsymbol{X}) = DecoderWithAttention(\boldsymbol{h}, y\textsubscript{n-1}, c_n)\label{decodereq}
\end{equation}

\begin{equation}
c\textsubscript{n} = Concat(softmax(QK^T)V, h_n)W_p\label{atteq}
\end{equation}

\vspace{10pt}

Where $c\textsubscript{n}$ is $n\textsuperscript{th}$ context vector, the key $K = \boldsymbol{h}W\textsubscript{k}$, value $V = \boldsymbol{h}W\textsubscript{v}$, and 
query $Q = h\textsubscript{n}W\textsubscript{q}$ such that W\textsubscript{q} , W\textsubscript{k}, W\textsubscript{v}, and W\textsubscript{p} are weight projection matrices $W\textsubscript{q} \in \mathbb{R}\textsuperscript{d, h\textsubscript{size}}$, $W\textsubscript{k} \in \mathbb{R}\textsuperscript{d, h\textsubscript{size}}$, $W\textsubscript{v} \in \mathbb{R}\textsuperscript{d, h\textsubscript{size}}$, and $W\textsubscript{p} \in \mathbb{R}\textsuperscript{h\textsubscript{size}, d+h\textsubscript{size}}$, where d is the projection size.

\begin{figure}[htb!]
  \includegraphics[width=\linewidth]{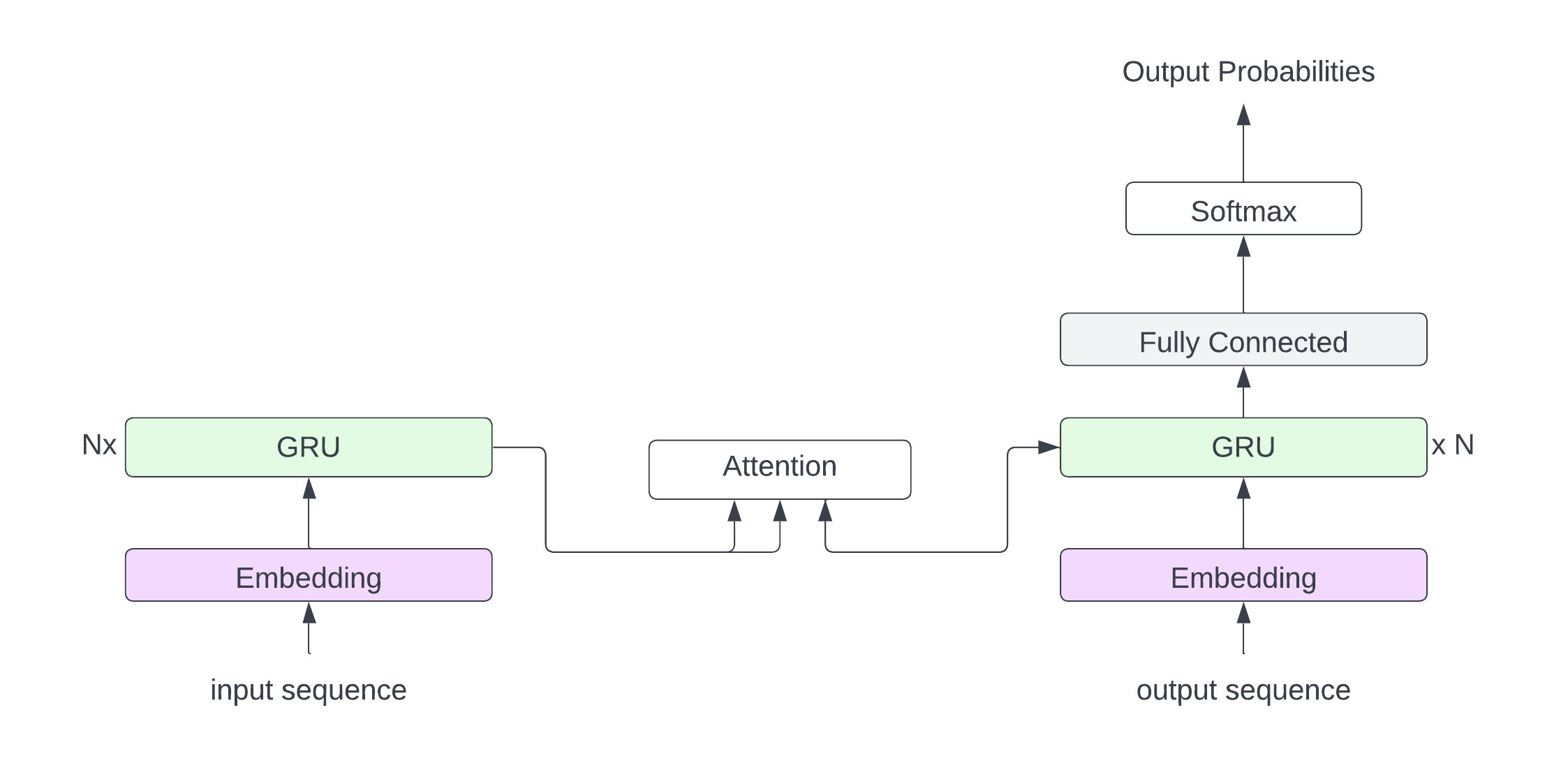}
  \caption{Attentional vanilla Seq2Seq using RNN.}\label{fig:vrnnmodel}
\end{figure}

Figure \ref{fig:vrnnmodel} visualizes the Attentional vanilla Seq2Seq using RNN model.

\subsubsection{Attentional Seq2Seq with stacked RNN blocks}

During experiments, we found that the vanilla Seq2Seq model described in \ref{vanilla} is highly affected by the corruption ratio. As the corruption increases, the model convergence becomes harder and slower and the model struggles to find the proper alignment so the attention collapse, which led to weak performance. We solved this issues by employing encoder-decoder using RNN blocks with attention instead of using stacked RNNs, where each RNN block consist of Gated Recurrent Unit (GRU) layer followed by dropout, feed-forward module, and layer normalization. 

\begin{figure}[htb!]
  \includegraphics[width=\linewidth]{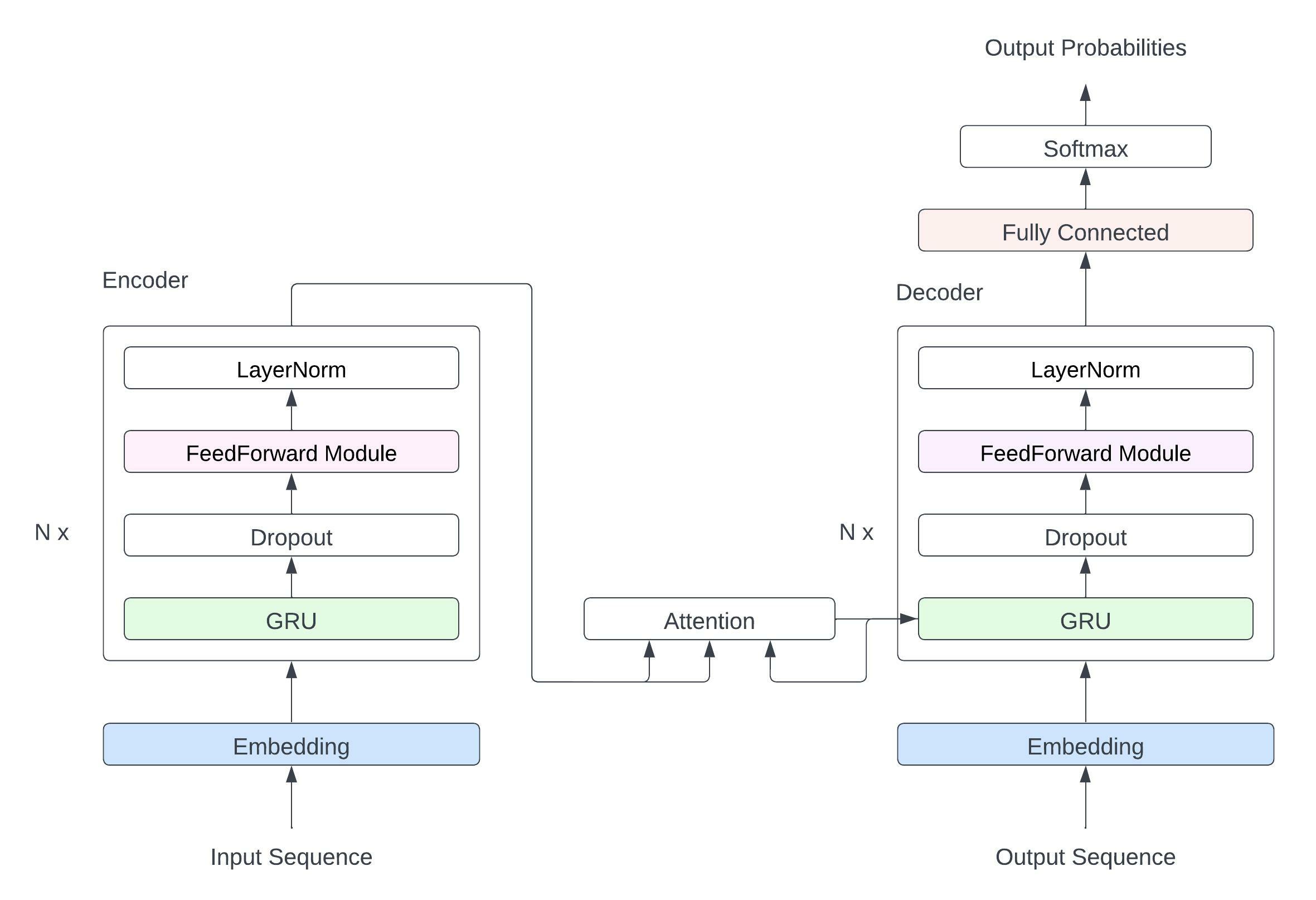}
  \caption{Attentional Seq2Seq with stacked RNN blocks.}\label{fig:rnnmodel}
\end{figure}

The feed-forward module consists of 2 fully connected layers such that the first up-scale the feature space by a factor of 2 and the second one scales it back to the original feature space dimension.
The encoder and the decoder with attention work the same way as illustrated in \eqref{encodereq}, \eqref{decodereq}, and \eqref{atteq}. Figure \ref{fig:rnnmodel} visualizes the Attentional Seq2Seq with stacked RNN blocks model.

\subsubsection{Transformer}
Transformer \cite{transformer} has become the go-to architecture in most of the NLP applications. Due its success in NLP, it has also been adopted in Computer Vision (CV) \cite{cv1, cv2}, Automatic Speech Recognition (ASR) \cite{sr1, sr2}, Speech Synthesis \cite{st1, st2}, among many others.

The transformer consists of an encoder where the input $\boldsymbol{X}$ gets mapped into high-level features $\boldsymbol{h}$, and a decoder with multi-head-attention that attends on $\boldsymbol{h}$ given the previously predicted samples $\boldsymbol{y\textsubscript{<n}}$.

The most notable advantage of Transformers over RNNs is that RNNs are auto-regressive in nature, which makes it difficult to take full advantage of modern fast computing devices such as Tensor Processing Units (TPUs) and Graphical Processing Units (GPUs). Another advantage stems from the limitation of the RNNs when encountering a very long sequence.

We used the model in its original architecture as mentioned in \cite{transformer} except for the feed-forward module, instead of up-scaling the feature space we downscale it by a  factor of 2.

\section{Experiments}

We have conducted a wide range of experiments, starting from how the performance changes while the error injection percentage changes to how different model architectures perform under the same error injection rate, taking the vanilla RNN model \ref{vanilla} as the baseline and improving based on it.

To test the models' performance, we used Character Error Rate (CER), Word Error Rate (WER), Character Error Reduction Rate (CERR), and Word Error Reduction Rate (WERR):

\begin{equation}
WER = \frac{S + D + I}{N}\label{wer}
\end{equation}

\begin{equation}
ERR = \frac{\overline{ER} - \widehat{ER}}{\overline{ER}}\label{werr}
\end{equation}

\vspace{5pt}

Where $\boldsymbol{S}$ is the number of substitutions, $\boldsymbol{D}$ is the number of deletions, $\boldsymbol{I}$ is the number of insertions, $\boldsymbol{N}$ is the number of words in the reference, $\boldsymbol{ERR}$ is the error reduction rate, $\boldsymbol{\overline{ER}}$ is the original/reference CER/WER, and $\boldsymbol{\widehat{ER}}$ is the CER/WER after correction.
CER can be calculated the same as in \eqref{wer}, but the main difference is that it works on the character level.

\subsection{Experimental Setup}

Data sets of different corruption ratios $\psi$ have been generated. Mainly 5\%, 10\%, 5\% and 10\% combined, and varied ratios between 2.5\% and 10\% are used as described in \ref{datageneration}.
For the Vanilla RNN model, we used 4 layers, a hidden size of 256, and an embedding size of 512. 
For the attentional Seq2Seq with stacked RNN blocks model, we used 3 layers, a hidden size of 256, and an embedding size of 512.
For both models, during the training process, we used gradient clipping with 1.0 max norm of the gradients and Adam optimizer with exponential learning rate (lr), decay \eqref{lr} with an initial learning rate of 10\textsuperscript{-4} and decay rate of 15*10\textsuperscript{-4}.

\begin{equation}
    lr(step) = \text{Initial learning rate} * (1 - \frac{\text{Decay rate}}{100})^\text{step} \label{lr}
\end{equation}

%\vspace{5pt}

For the transformer model we used four layers, 512 model dimensionality and 8 heads. For training, Adam optimizer is used with learning rate scheduler as mentioned in \cite{transformer} with 4000 warm-up steps.

We trained our models on one machine with two NVIDIA 3080 TI GPUs, and we used Kullback–Leibler divergence loss for all models.
During training, we employed different regularization techniques. We used dropout \cite{dropout} with a 10\% dropout ratio and label smoothing of value $\epsilon$=0.1 \cite{labelsmoothing}. 
During the training process of the vanilla RNN model, the model struggled to build an alignment when the corruption ratio increased from $\psi$=5\% to $\psi$=10\%, and it took 20X times to converge and build alignment. To achieve that, we used a simple trick to speed the process up by pre-training the model on low corruption ratio ($\psi$=5\%) for few steps till the model learns to build a proper alignment, then we retrained it on the data of high corruption ratio.

\subsection{Results}

A test set of 100K sentences have been used. In each experiment, we tested the WER, CER, WERR, CERR on $\psi$=5\% and $\psi$=10\%.
Table \ref{table:result} shows the CER and WER results, while table \ref{table:original} shows the original CER and WER calculated on the original corrupted test data. 
Lastly, Table \ref{table:reductionrates} shows the WERR, and CERR. We can see that the transformer model outperforms all other models under all data conditions. From tables \ref{table:result} and \ref{table:reductionrates} we can see that adding more data would help the performance, as mixing the 5\%, and the 10\% data together yields better performance across all models.

\vspace{-10pt}

\begin{table}[htb!]
\centering
  \begin{tabular}{lcccccc}
    \toprule
    \multirow{2}{*}{Model} &
      \multicolumn{2}{c}{CER (\%)} &
      \multicolumn{2}{c}{WER (\%)} \\
      & {$\psi=5\%$} & {$\psi=10\%$} & {$\psi=5\%$} & {$\psi=10\%$} \\
      \midrule
    Transformer\_0.05 & 1.24 & 4.15 & 5.35 & 18.38 \\
    Transformer\_0.1 & 1.45 & 2.82 & 5.95 & $\boldsymbol{10.36}$ \\
    Transformer\_mixed & $\boldsymbol{1.11}$ & $\boldsymbol{2.8}$ & $\boldsymbol{4.8}$ & 10.65 \\
    Transformer\_varied & 1.22 & 3.16 & 5.41 & 12.35 \\
    RNNB\_0.05 & 1.74 & 4.45 & 7.76 & 18.8 \\
    RNNB\_0.1 & 1.86 & 4.01 & 7.8 & 15.41 \\
    RNNB\_mixed & 1.67 & 4.06 & 7.51 & 16.55 \\
    RNNB\_varied & 1.77 & 4.36 & 8.14 & 18.01 \\
    VanillaRNN\_0.05 & 1.89 & 4.99 & 8.33 & 20.67 \\
    VanillaRNN\_0.1 & 2.01 & 4.52 & 18.19 & 16.95 \\
    VanillaRNN\_mixed & 1.86 & 4.53 & 7.84 & 17.36 \\
    VanillaRNN\_varied & 2.01 & 4.97 & 8.76 & 19.49 \\
    \bottomrule
  \end{tabular}
  \caption{CER and WER tested on 5\% and 10\% corruption ratios, for a model Y\_X, Y is the model name and X is the corruption ratio that Y is trained on.}
  \label{table:result}
\end{table}

\vspace{-25pt}

\begin{table}[htb!]
\centering
  \begin{tabular}{lcc}
    %\toprule
      $\psi$ (\%) &
      $\overline{CER}$ (\%) &
      $\overline{WER}$ (\%) \\
      \hline
    10 & 10.02 & 50.94 \\
    5 & 5.03 & 29.72 \\
    \bottomrule
  \end{tabular}
  \caption{CER and WER on the test set after introducing data corruption.}
  \label{table:original}
\end{table}

\begin{table}[htb!]
\centering
  \begin{tabular}{lcccccc}
    \toprule
    \multirow{2}{*}{Model} &
      \multicolumn{2}{c}{CERR (\%)} &
      \multicolumn{2}{c}{WERR (\%)} \\
      & {$\psi=5\%$} & {$\psi=10\%$} & {$\psi=5\%$} & {$\psi=10\%$} \\
      \midrule
    Transformer\_0.05 & 75.34 & 58.58 & 81.99 & 63.91 \\
    Transformer\_0.1 & 71.17 & 71.85 & 79.97 & $\boldsymbol{79.66}$ \\
    Transformer\_mixed & $\boldsymbol{77.93}$ & $\boldsymbol{72.05}$ & $\boldsymbol{83.84}$ & 79.09 \\
    Transformer\_varied & 75.74 & 68.46 & 81.79 & 75.75 \\
    RNNB\_0.05 & 65.40 & 55.58 & 73.88 & 63.09 \\
    RNNB\_0.1 & 63.02 & 59.98 & 73.75 & 69.74 \\
    RNNB\_mixed & 66.79 & 59.48 & 74.73 & 67.51 \\
    RNNB\_varied & 64.81 & 56.48 & 72.61 & 64.64 \\
    VanillaRNN\_0.05 & 62.42 & 50.19 & 71.97 & 59.42 \\
    VanillaRNN\_0.1 & 60.03 & 54.89 & 38.79 & 66.72 \\
    VanillaRNN\_mixed & 63.02 & 54.79 & 73.62 & 65.92 \\
    VanillaRNN\_varied & 60.03 & 50.39 & 70.52 & 61.73 \\
    \bottomrule
  \end{tabular}
  \caption{CERR and WERR tested on 5\% \& 10\% corruption ratios, for a model Y\_X, Y is the model name and X is the corruption ratio that Y is trained on.}
  \label{table:reductionrates}
\end{table}

\begin{figure}[htb!]
  \includegraphics[width=0.95\textwidth]{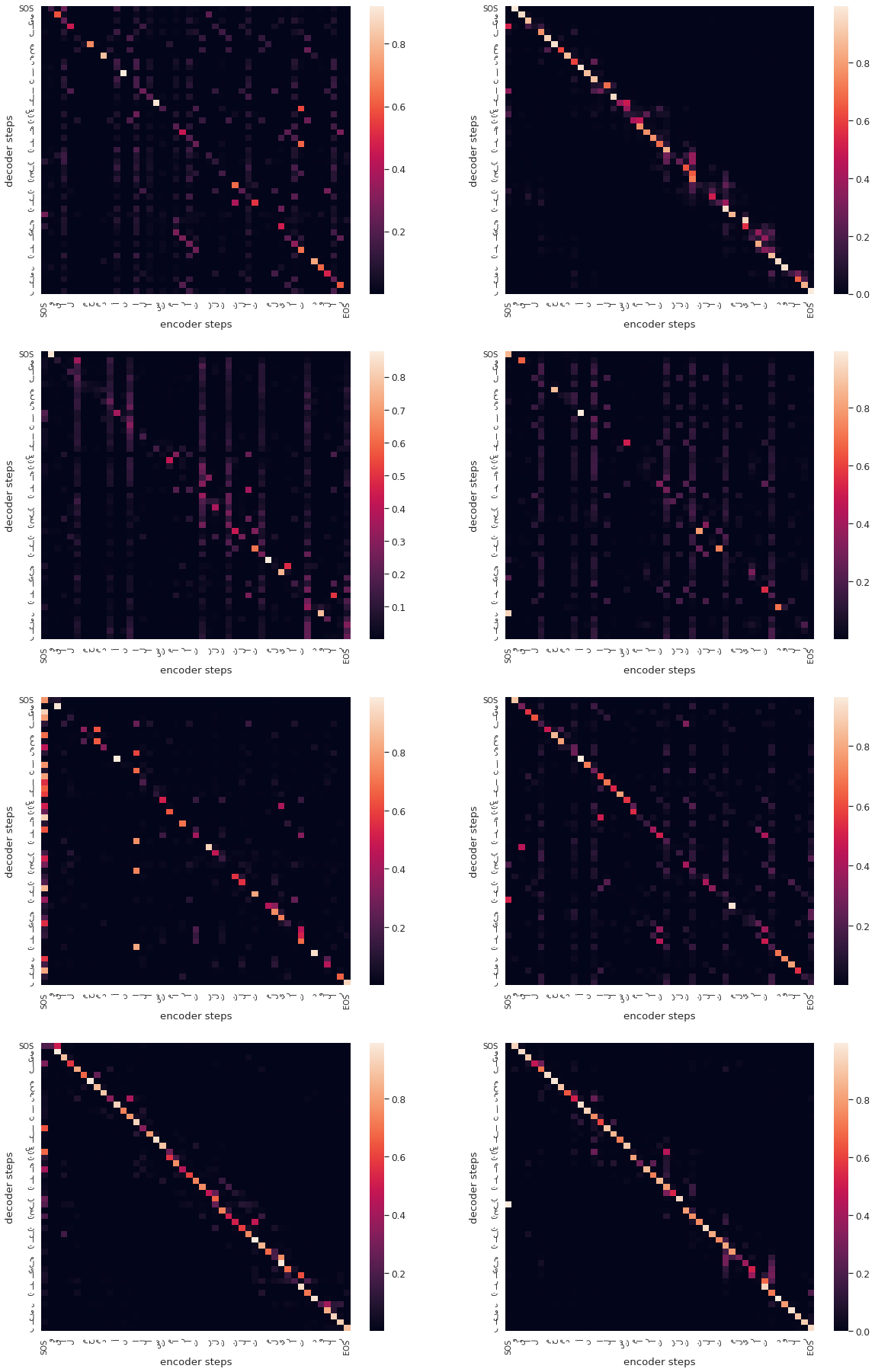}
  \caption{Attention generated during inference across all heads of the last decoder layer of the transformer model.}\label{fig:attention}
\end{figure} 

\begin{table}[htb!]
\centering
\setcode{utf8}
\begin{tabularx}{0.95\textwidth} { 
  | >{\centering\arraybackslash}X 
  | >{\centering\arraybackslash}X| }
 \hline
 Input & Result \\
 \hline
\RL{إن الذب من المعادرن النفيسة} &
\RL{إن الذهب من المعادن النفيسة} \\
 \RL{وقال محمد أن الاستمارات بلت ثلاث مليرات دولار}   &
 \RL{وقال محمد أن الاستثمارات بلغت ثلاث مليارات دولار}   \\
 \RL{ذبه الرجل مسراعا إلى الابة} &
 \RL{ذهب الرجل مسرعا إلى الغابة} \\
\RL{إلى أيييين أنت ناظؤ} &
\RL{إلى أين أنت ناظر} \\
\RL{الأؤدن عامصتها عمان} &
\RL{الأردن عاصمتها عمان} \\
\RL{هل سبق ورزت الدينة} &
\RL{هل سبق ورزت المدينة} \\
% \RL{تحتاج النتات إلى مالء} &
% \RL{ تحتاج النباتات إلى ماء }  \\
\RL{يخج من البكرات مواد مصرة وغازات} &
\RL{يخرج من البكرات مواد مصرية وغازات}\\ 
% \RL{اليمهو الأحد وبعد الغد هو الثلثاء} &
% \RL{اليوم هو الأحد وبعد الغد هو الثلاثاء} \\
\RL{وثد كنا نرتدي اليز الرمسي} &
\RL{وقد كلنا نرتدي الزي الرسمي}\\
\RL{الناق الذي يحط بسحل المحط الهاي} &
\RL{النطاق الذي يحيط بسواحل المحيط الهادي} \\
\RL{يختلف شكل المخط البركاني باختاف الماد الي يترلاكب منها}&
\RL{يختلف شكل المخطط البركاني باختلاف المواد التي يتراكب منها} \\
\hline
\end{tabularx}
  \caption{Samples generated during inference from the transformer model.}\label{table:samples}
\end{table}

Figure \ref{fig:attention} shows the attention generated from the transformer model trained on the mixed data set across different heads during inference, and table \ref{table:samples} shows testing samples during inference.

\newpage

\section{Concluding Remarks}

We presented "AraSpell," a framework for Arabic spelling correction, and introduced different Seq2Seq models with error injection schema. Our model was trained on more than 6.9 million sentences and it has achieved a Character Error Rate (CER) of 1.11\%, and Word Error Rate (WER) of 4.8\%, which resulted in 77.93\%, and 83.84\% character and word error reduction rate respectively. Moreover, the proposed model also achieved 2.9\% CER and 10.65\% WER resulting in 72.05\%, and 79.09\% character and word error reduction rate, respectively. Both results achieved on a test set of 100K sentences with 5\%, and 10\% error/corruption injection rate.

A main feature of the proposed approach is the proper use of error injection. For further improvements, in future work, we recommend adding more grammatical mistakes for error injection and training on longer sentences to capture more context. The approach is promising and can be put in practical use, especially if further training is applied.

\section*{Authors’ Contributions} 

Development, implementation and experimental analysis: MS. Writing and reviewing: MS and FNA. Supervision: FNA.

\section*{Supplementary Materials} 

Software and data produced as part of this study are available at the following github repository: \url{https://github.com/msalhab96/AraSpell}

\section*{Funding and/or Conflicts of interests/Competing interests.}

No funding was received for conducting this study.
Moreover, 
the authors have no competing interests or conflict of interest to declare that are relevant to the content of this article.

%\section*{Declarations}

%The authors did not receive support from any organization for this work. The authors have no relevant financial or non-financial interests to disclose.

%\bibliographystyle{sn-mathphys}
\bibliography{references}
\end{document}